\pdfoutput=1

\documentclass[conference]{IEEEtran}
\usepackage[pdftex]{graphicx}   
\usepackage{amsmath}
\usepackage{algorithm}
\usepackage{algorithmic}

\begin{document}
%
\title{Causal Inference via Conditional Kolmogorov Complexity using MDL Binning}

\author{\IEEEauthorblockN{Daniel Goldfarb}
\IEEEauthorblockA{Northeastern University\\
\& GE Global Research\\
goldfarb.d@husky.neu.edu}
\and
\IEEEauthorblockN{Scott Evans}
\IEEEauthorblockA{GE Global Research\\
evans@ge.com}}

\maketitle

\begin{abstract}
Recent developments have linked causal inference with Algorithmic Information Theory, and methods have been developed that utilize Conditional Kolmogorov Complexity to determine causation between two random variables. We present a method for inferring causal direction between continuous variables by using an MDL Binning technique for data discretization and complexity calculation. Our method captures the shape of the data and uses it to determine which variable has more information about the other. Its high predictive performance and robustness is shown on several real world use cases. \\

\end{abstract}


%
\IEEEpeerreviewmaketitle

\section{Introduction}
Kolmogorov Complexity is the length of the shortest binary program to create a given string X and measures the descriptive complexity of individual sets of data or probability distributions. The Conditional Kolmogorov Complexity, $K(X | Y)$, is the size of the smallest program required to create string $X$ given input $Y$. As described in \cite{janzing2010} \cite{budha2016} \cite{budha2017}, when $K(X)+K(Y|X) < K(Y)+K(X|Y)$ we can conclude $X \rightarrow Y$ ($X$ causes $Y$). The challenge is that Kolmogorov complexity is incomputable due to the Halting problem. Methods such as compression techniques and Stochastic Complexity \cite{barron1998, budha2017} have been developed to estimate $K(X)$ and $K(X|Y)$. We will map feature data and their corresponding probability distributions to binary strings to determine causal features in data by estimating Kolmogorov Complexity and Conditional Kolmogorov Complexity of these strings. \cite{janzing2010} \cite{budha2016} \cite{budha2017} have several approaches that will be leveraged. We will go beyond these approaches by using an MDL Binning technique to discretize continuous data and treat the binning techniques as the model in terms of the MDL principle. The estimated complexity is then the cost of describing random variable $X$ using a given binning technique, and $K(X)$ is approximated as the minimum complexity of $X$ over all binning techniques.

\vspace{0.4cm}

\noindent
The main contributions of our work are as follows:
\begin{itemize}
\item We present a new MDL binning technique to provide estimates of Kolmogorov Complexity of continuous data as a discrete probability distribution.
\item We develop a method for using MDL binned distributions to determine conditional $K(X|Y)$.
\item We show how this approach gives improved robustness in determining causal direction over state-of-the-art techniques.
\end{itemize}

\section{Prior Work}

\subsection{Causal Inference via Algorithmic Information Theory}
Budhathoki et al. used a couple methods to infer causality using algorithmic information theory \cite{budha2016} \cite{vreeken2017}. He inferred the most likely causal direction between random variables by identifying lowest K-complexity. If $K(X) + K(Y|X) < K(Y) + K(X|Y)$ then $X \rightarrow Y$. These authors used a tree packing algorithm to compress binary data to compute the complexities using the MDL principle. Since the packing algorithm does not support the compression of non-binary data off-the-shelf, binarization of the data is required as a pre-processing step. Marx and Vreeken \cite{marx2017} developed a similar causal inference algorithm that uses the same inference for predicting causal direction. Their method uses regression to compress the data by encoding functional relations which allows for the ability to make causal inference on continuous data.


\subsection{The Minimum Description Length Principle in Coding and Modeling}
Barron et al. outlined the principles of MDL in a handful of applications for data compression and statistical modeling \cite{barron1998}. One of these applications is Density Estimation which utilizes a histogram density function to assign points to bins. Our calculation of the tradeoff between model cost and error cost falls closely with the principles in this application. We extended this method to iteratively calculate and log complexities for a variety of bin numbers to determine the minimal Kolmogorov complexity estimation.

\vspace{0.7cm}
\section{Encoding a Distribution to Estimate Kolmogorov Complexity}

The essence of computing the complexity of a continuous random variable is in how we discretize it via the binning technique. Hence we will briefly outline our two proposed techniques before describing our complexity estimation algorithm. After that we will continue with the analysis of the binning techniques by comparing their performance in concise visual plots.

\subsection{Definitions of Binning Techniques}
In order to compute complexities we must first find a binning assignment that is simple enough that it doesn't cost too much but still captures the essence of the distribution:
\begin{itemize}
\item Uniform: assign equal sized bins to span the range of points
\item Greedy: iteratively add variable sized bins to minimize complexity
\end{itemize}

These binning techniques are performed iteratively over number of bins in order to find the best binning strategy for the given sampling of points. Once the optimal technique is found, the complexity of the distribution is defined as the Kolmogorov complexity estimation given those optimal bins. 

\subsection{Computing Kolmogorov complexity for sampled distribution $X: K(X)$}

\noindent
Our method for estimating the complexity of a random variable is as follows:

\begin{algorithm}
\caption{Calculate $Complexity(X, bins)$}
\begin{algorithmic} 
\STATE{Initialize $complexity=0$}
\FOR{$b \in bins$} 
\STATE{Calculate and store Shannon code for $b$}
\STATE{complexity += 1 + $log_2(len($Shannon code for $b))$}
\ENDFOR
\FOR{$p \in X$}
\STATE{$cl$ = code length of $bin_p$}
\STATE{$mean$ = avg of points in $bin_p$}
\STATE{complexity += $log_2(cl)+log_2(|p-mean|))$}
\ENDFOR
\RETURN $complexity$
\end{algorithmic}
\end{algorithm}

\begin{algorithm}
\caption{Calculate $K(X)$}
\begin{algorithmic}
\STATE{Initialize $best\_complexity=\infty$}
\FOR{$B \in$ set of possible binning strategies$)$}
\IF{$Complexity(X, B) < best\_complexity$}
\STATE{$best\_complexity = Complexity(X, B)$}
\ENDIF
\ENDFOR
\RETURN{$best\_complexity$}
\end{algorithmic}
\end{algorithm}

\vspace{0.4cm}

\noindent
\begin{equation*}
\begin{gathered}
\hspace{-2.7cm} K(X):=\#bins + \Sigma_{b \in bins}log(CL(b))+\\
\hspace{1cm} \Sigma_{p \in X}[log(CL(bin_p))+log(|p-mean(bin_p)|)]
\end{gathered}
\end{equation*}

\vspace{0.5cm}

\noindent
We see that the calculation of complexities is split into three parts, defined as follows: \\
\vspace{-0.4cm}
\begin{itemize}
\item \textbf{Model Cost:} Total number of bins and length of Shannon code for each bin.
\item \textbf{Code Length Cost:} Length of bin Shannon code for each point in $X$.
\item \textbf{Error Cost:} Difference between each point's value and the mean of all points in its bin.
\end{itemize}

\vspace{0.3cm}

\noindent
The tradeoff, via the MDL principle, between these three components is explored in the next section.

\vspace{0.5cm}

\subsection{Comparing Uniform vs. Greedy Binning Techniques}

As seen in \textit{Figure 2}, the greedy method finds a local optimum earlier than the uniform method, but over time the best global optimum is found by the uniform method. The greedy decisions made early on are not beneficial for binning in the long run. The dataset used in this example is a 1,000 point bimodal normal distribution with a $40/60$ spread, but the same sentiment follows with toy and use case datasets used.

\begin{center}
\includegraphics[scale=0.5]{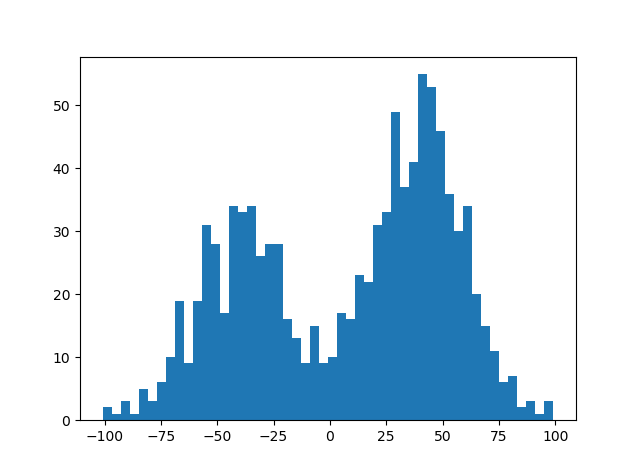}\\
\textit{Figure 1: Toy Bimodal Distribution}
\end{center}

\begin{center}
\includegraphics[scale=1.2]{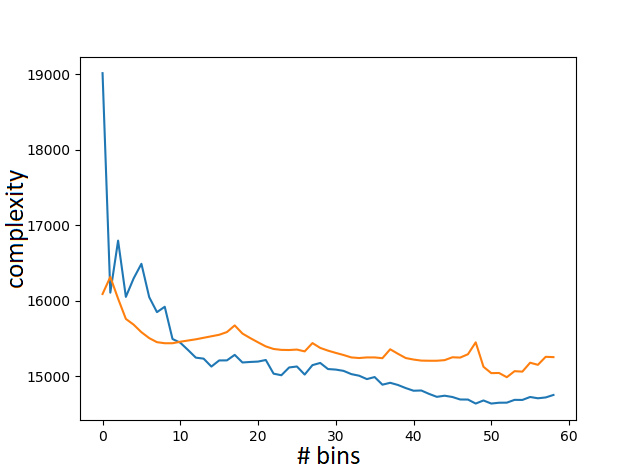}\\
\includegraphics[scale=0.2]{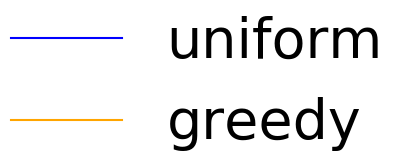}\\
\textit{Figure 2: Comparing the optimization of Uniform and Greedy binning techniques}
\end{center}
\vspace{0.4cm}

\vspace{0.4cm}

Since for the uniform method, we are trying one binning for each of the $n$ bin options and computing K-complexity takes linear time, the runtime is $O(n^2)$. However, the greedy method additionally has to test $m$ possible partitions for each step, which results in a runtime of $O(n^2m)$. For both of these reasons, we will be focusing on the uniform method from now on.

\vspace{-0.3cm}

\begin{center}
\includegraphics[scale=1.2]{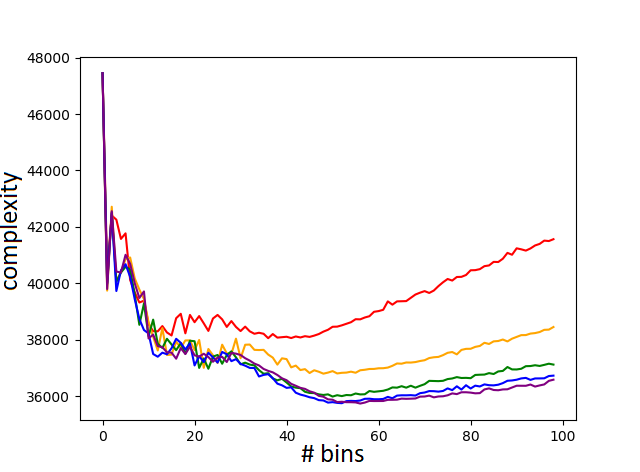}\\
\textit{Figure 3: Comparing the optimization of the Uniform binning technique on five different normal distribution sampling sizes}
\end{center}
\vspace{0.4cm}

\textit{Figure 3} shows the uniform binning complexity plots for 500, 1,000, 1,500, 2,000, and 2,500 point normal samplings from top to bottom. We find that for the 1,000-2,500 point samplings, the uniform binning technique produces relatively similar optimal bin numbers ($\sim$55). However for the 500 point distribution (in red), not enough points were sampled which resulted in a simpler representation of 20 uniform bins. We conclude that at least 1,000 points is enough to properly represent this distribution and larger sample sizes have little effect on the optimal binning strategy. We use this knowledge moving forward by narrowing our use cases to only those with $>500$ rows.

\begin{center}
\includegraphics[scale=1.2]{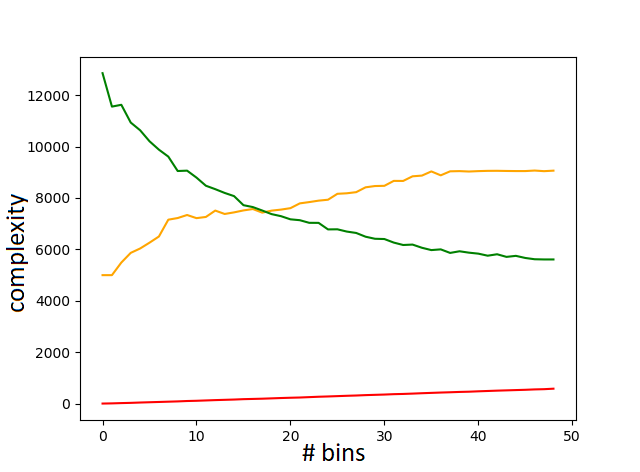}\\
\includegraphics[scale=0.3]{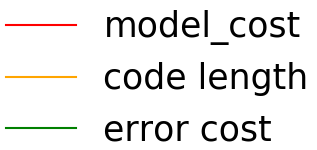}\\
\textit{Figure 4: Tradeoff between cost components for the normal distribution}
\end{center}

\begin{center}
\includegraphics[scale=1.2]{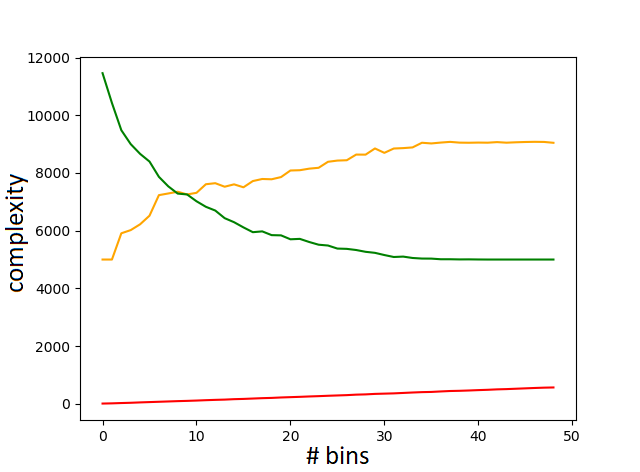}\\
\includegraphics[scale=0.3]{"plot_legend".png}\\
\textit{Figure 5: Tradeoff between cost components for the skew distribution}
\end{center}

\begin{center}
\includegraphics[scale=1.2]{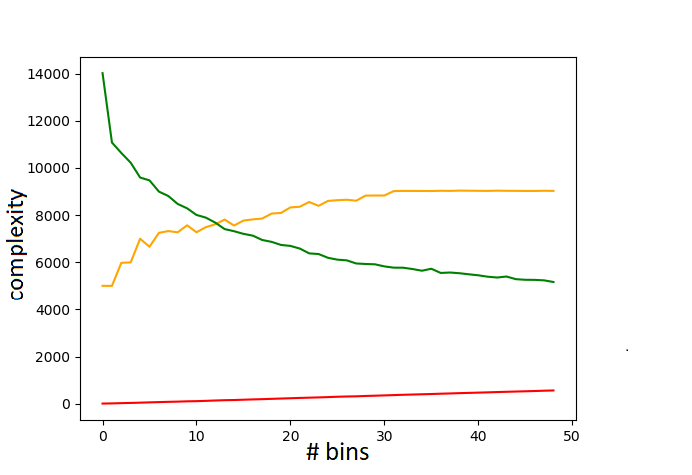}\\
\includegraphics[scale=0.3]{"plot_legend".png}\\
\textit{Figure 6: Tradeoff between cost components for the bimodal distribution}
\end{center}

Computing optimal complexities via the MDL principle poses a tradeoff between model cost, code length cost, and error cost. We see from various types of probability distributions in \textit{Figures 4, 5, 6} that as we increase the number uniform bins, the model cost increases to store information about the type of binning technique used, the code length cost increases in order to represent the code lengths of a larger number of bins, and the error cost decreases since there is a lower average difference between values of points and the mean of their respective bins. As defined in subsection $B$, the sum of these three values are summed up and kept track of at every binning iteration. The minimum of sums is defined as the estimated Kolmogorov complexity, with optimal bins being calculated by its corresponding uniform bin number.

\vspace{0.4cm}

\section{Inferring Causality}

Recall that if $K(X)+K(Y|X) < K(Y)+K(X|Y)$ then it is most likely that $X \rightarrow Y$. We already defined how to estimate $K(X)$ and $K(Y)$ so what's left is to compute $K(X|Y)$ and $K(Y|X)$. We find these conditional complexities in an analogous way to the original ones, however we take binning information from the conditional variables for free without incorporating the corresponding model cost.

\subsection{Computing Kolmogorov complexity for $X$ given $Y: K(X|Y)$}

\noindent
Our method for estimating the conditional complexity of one random variable $X$ given another random variable $Y$ is as follows:

\begin{algorithm}
\caption{Calculate $K(X|Y)$}
\begin{algorithmic} 
\STATE{Initialize $complexity = 0$}
\STATE{$bins_Y$ = optimal uniform bins for $Y$}
\STATE{$bins_X$ = $len(bins_Y)$ bins scaled for $X$}
\STATE{$counts_X$ = bin counts for $bins_X$}
\STATE{$counts_Y$ = bin counts for $bins_Y$}
\WHILE{$\exists i,j$ such that $counts_Y[i]>counts_X[i]$ and $counts_Y[j]<counts_X[j]$}
\STATE{$diff\_i = |counts_Y[i]-counts_X[i]|$}
\STATE{$diff\_j = |counts_Y[j]-counts_X[j]|$}
\STATE{$counts\_diff = min(diff\_i, diff\_j)$}
\STATE{$counts_X[i]$ += $counts\_diff$}
\STATE{$counts_X[j]$ -= $counts\_diff$}
\STATE{$complexity$ += $1+log_2(len(bins_X))$}
\ENDWHILE
\STATE{$shannon_X$ = Shannon codes for $counts_X$}
\FOR{$p \in X$}
\STATE{$mean$ = avg of points in $bins_X[p]$}
\STATE{complexity += $log_2(shannon_X[p])+log_2(|p-mean|))$}
\ENDFOR
\RETURN $complexity$
\end{algorithmic}
\end{algorithm}

\vspace{0.4cm}

\noindent
\begin{equation*}
\begin{gathered}
\hspace{-0.2cm} K(Y|X):=\#binsbalanced + \Sigma_{b \in bins} I(balanced)[log(CL(b))\\
\hspace{-0.3cm} +log(ind(b))]+\Sigma_{p \in X}[log(CL(bin_p))\\
\hspace{-2.2cm} +log(|p-mean(bin_p)|)]
\end{gathered}
\end{equation*}

\vspace{0.4cm}

\noindent
The difference here is that we use the optimal bin number for $Y$ to compute $K(X)$ and then iteratively balance out code lengths of bins to better fit the distribution of $X$. The penalty for balancing out each code length is an additional model cost encoding along with the bin indices of both edited counts.

\vspace{0.4cm}

In some cases, we find that $K(X|Y)>K(X)$ due to additional cost of indicating the indices of the new balanced bins. To follow probabilistic axioms, we define our Kolmogorov complexity estimate to be $min(K(X|Y),K(X))$ to follow that if $Y$ gives no information about $X$, then $K(X|Y):=K(X)$.

For our inferences we will no longer incorporate error cost since we find that it is commonly uniformly distributed from bin to bin since they are always being compared to bin means. However, the error cost is still used when performing the tradeoff to find the optimal binning strategy. This also allows the model cost to play a bigger role in determining overall complexity which is important for our bin balancing algorithm. After we lay out some prior work to provide a better description of the area, we will be using this method on several real world use cases and one toy example.

\section{Data Selected and Use Case}

Our real world use cases are all datasets that are accessible from a handful of sources [UCI Machine Learning Repository, Econometrics Toolbox by James P. Lesage, $\&$ Kaggle] and took inspiration from two papers \cite{vreeken2017} \cite{mooij2016}. The first is the ORIGO paper from Budhathoki et al. This paper lays out a similar algorithmic information theory method for discrete data, so we made sure to use their use cases for comparison purposes. The other paper by Mooij et al. uses a less related method for inferring cause and effect but provides $\sim100$ different variable pair examples with intuitive ground truth from which we extracted the use cases that made sense for our method. Namely, those with $>500$ rows and preferrably no binary features. We also include a toy solar power use case which was the industrial application that motivated us to perform causal inference on continuous data.

It should be noted that each of the datasets are normalized to have minimum value 0, maximum value 100, and the rest of the points scaled accordingly. This detail makes sure that when computing the optimal bin strategy, error costs are bounded below by $1$ and above by $ceil(log_2(100))=7$. This bounding of error costs allows the scaling of our model to datasets with extremely large values so that the error costs do not overrule the model costs and code length costs when performing the iterative binning complexity tradeoff.

\textit{Table 1} provides a summary of the results when running our causal inference technique on the use cases. Both sides of the inequality are given along with the computed percent change between $K(Y)+K(X|Y)$ and $K(X)+K(Y|X)$ to provide the likelihood of causality. Namely, since the inequality holding implies that it is \textit{most likely} that $X$ causes $Y$, then a larger magnitude in difference implies a greater likelihood. In order to normalize over datasets with many points, we provide this likelihood metric in percent change.

Out of 11 pairwise causal examples, our method predicted 8 of them to be causal, 2 of them to be noncausal, and 1 of them as inconclusive. We came to the inconclusive result because $K(Y|X)>K(Y)$ and $K(X|Y)>K(X)$, so our evaluation found $K(Y)+K(X|Y)=K(X)+K(Y|X)$. So our precision score for this probing of examples is $80\%$ which is on-par with other continuous causal inference methods \cite{janzing2012} \cite{peters2014}. This result is using a conclusivity threshold of 0 meaning that any positive percent change will predict \textit{causal}, any positive percent change will predict \textit{not causal}, and a 0 percent change will predict \textit{inconclusive}. In addition to the high precision, we find that 

\vspace{0.4cm}

\begin{table}[hp]
{\normalsize
	\begin{tabular}{|l|l|l|l|l|l|l|}
	\hline
	dataset        & $X$          & $Y$          & $K(Y)+ K(X|Y)$ & $K(X)+ K(Y|X)$ & result & \% change \\ \hline
	car evaluation & safety     & evaluation & 24221       & 22527  &   $X \rightarrow Y$  & 7.5       \\ \hline
	abalone        & sex        & length     & 34939       & 28366    & $X \rightarrow Y$    & 23.2      \\ \hline
	abalone        & sex        & diameter   & 35368       & 28366   & $X \rightarrow Y$    & 24.7      \\ \hline
	abalone        & sex        & height     & 29226       & 28366   & $X \rightarrow Y$    & 3.0       \\ \hline
	adult            & education  & income & 348980   & 384909 & $Y\rightarrow X$ & -9.3 \\ \hline
	concrete      & cement   & strength  & 15129 & 15129 & inconclusive & 0 \\ \hline
	concrete     & water    & strength    & 15013  & 15258 & $Y \rightarrow X$ & -1.6 \\ \hline
	concrete    & superplast & strength & 14374  & 14331  & $X \rightarrow Y$ & 0.3 \\ \hline
	county         & population & employment & 46261       & 46086   & $X \rightarrow Y$    & 0.3       \\ \hline
	housing      & rooms      & value       & 8061        & 7903  & $X \rightarrow Y$      & 2.0       \\ \hline
	toy solar      & solar      & power      & 15029       & 14680   & $X \rightarrow Y$    & 2.4       \\ \hline
	\end{tabular}
}
\end{table}

\begin{center}
\textit{Table 1: Causal Inference Summary Results}
\end{center}

\vspace{0.4cm}

\noindent
our inferences are robust with respect to the threshold used. \textit{Figure 7} shows that for conclusivity thresholds between $0\%$ and $5\%$, precision values stay at $80 \pm 6\%$.

\vspace{-0.2cm}

\begin{center}
\includegraphics[scale=1.2]{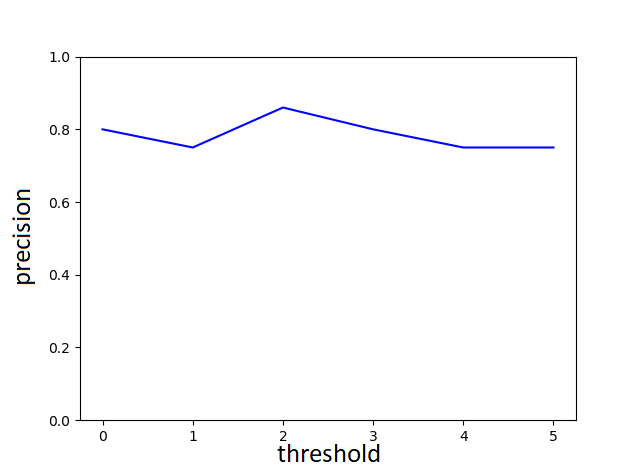}\\
\textit{Figure 7: Comparing the precisions of six causal inference classifiers to show the robustness of our method}
\end{center}

\vspace{0.4cm}

We will now dig into on some of the important use case examples and elaborate on the intuition behind their ground truths.

\subsection{Car Evaluation Dataset}
The car evaluation dataset contains 1728 rows of 6 features about used cars for sale. The labels to be predicted are the evaluation prices. The two features that we are isolating are the safety rating of the car and its evaluation label. We hypothesize that since safer cars have special technologies that they should cause higher prices. However, a higher price does not imply a safer car. Plenty of sports cars are expensive due to their fancy engines but are not necessarily safer, so this is a strictly causal and non-symmetric relationship. We were able to correctly infer this direction of causality.

\subsection{Abalone Dataset}
The abalone dataset contains 4177 rows of 8 features about different abalone shellfish. The labels to be predicted are the
\\\\\\\\\\\\\\\\\\\\\\\\\\\\\\\\\\ age of the abalone using number of rings on the shell as a proxy. The two features that we are isolating are the sex of the abalone and its size (length, diameter, height). It is common in other species that on average males are larger than females. We hypothesize that the same relationship holds for abalones as well. On the other hand, changing the size of an abalone to be larger does not make it more likely to be male, so this is a causal relationship. We were able to correctly infer this direction of causality.

\subsection{Housing Dataset}
The housing dataset contains 506 rows of 14 features about the details and neighborhoods of Boston apartments. The labels to be predicted are the values of the homes in $1,000$'s of dollars. The two features that we are isolating are the number of rooms in the apartment and the value. More rooms in an apartment adds value by providing more space and accomdation. However an increase in price does not imply more rooms in the apartment. For example, an apartment may be more expensive due to its location with respect to the city or the neighborhood it belongs to. So we infer that this is a causal relationship where number of rooms causes apartment value. We were able to correctly infer this direction of causality.

\subsection{Toy Solar Power Dataset}
This toy example is inspired by the industrial application of the input and output to solar panels. Each of the 1,000 points is an instance in time where the intensity and instantaneous value of power generation are logged. Given a normal distribution with a single mean and infrequent extreme intensities for $X$, the outputted power generation distribution, $Y$, is an extremely skewed normal distribution.

\vspace{1cm}

\begin{center}
\includegraphics[scale=0.4]{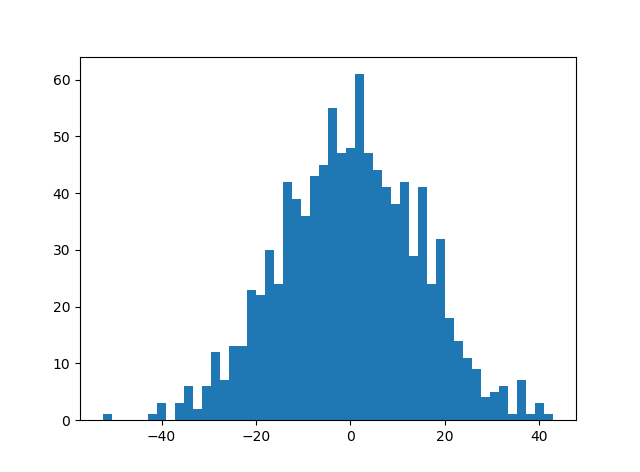}\\
\textit{Figure 8: Toy Solar Distribution}
\end{center}

\begin{center}
\includegraphics[scale=0.4]{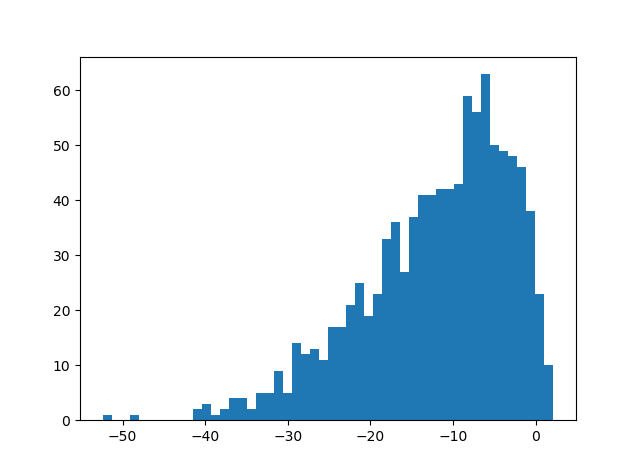}\\
\textit{Figure 9: Toy Power Distribution}
\end{center}

\vspace{0.4cm}

This shape follows from the clipping behavior of power generation. Solar panels are not perfect, they have a clipping point where any more solar intensity does not increase the marginal power. As a result, we see a plateauing behavior in the power and thus a hard clip in its probability distribution. So given a level of solar intensity, we should be able to tell what the level of power generation was at that point. However, this is not possible going from $Y$ to $X$. If we are given a point of power generation that is above the clipping line, it is impossible to recover what the level of solar intensity was at that instance. Hence, this is a causal relationship. We were able to correctly infer this direction of causality.

\section{Conclusion}
We have introduced a causal inference technique that uses the MDL binning principle to compress and compute Kolmogorov complexities for continuous data. We applied our method to numerous real world examples with intuitive ground truths and showed competitive prediction precision against state-of-the-art methods and robustness over various conclusivity thresholds.

For future work, we are interested in applying our pairwise causal inference method to feature selection. Given a dataset $X$ and set of labels $Y$, we want to extract a causal feature set such that the only features used in the new $X'$ have a causal relationship with $Y$. This in turn would produce a causal machine learning model for which we know that each feature is a causal predictor of $Y$.

\end{document}